\documentclass{article}
\usepackage{ijcai13}

\usepackage{times}
\usepackage{graphicx}
\usepackage[tight,footnotesize]{subfigure}
\title{Facial Expression Detection using Patch-based Eigen-face Isomap Networks}
\author{Sohini Roychowdhury \\
University of Washington, Bothell\\
WA, USA-98011 \\
roych@uw.edu}

\begin{document}

\maketitle

\begin{abstract}
Automated facial expression detection problem pose two primary challenges that include variations in expression and facial occlusions (glasses, beard, mustache or face covers). In this paper we introduce a novel automated patch creation technique that masks a particular region of interest in the face, followed by Eigen-value decomposition of the patched faces and generation of Isomaps to detect underlying clustering patterns among faces. The proposed masked Eigen-face based Isomap clustering technique achieves 75\% sensitivity and 66-73\% accuracy in classification of faces with occlusions and smiling faces in around 1 second per image. Also, betweenness centrality, Eigen centrality and maximum information flow can be used as network-based measures to identify the most significant training faces for expression classification tasks.  The proposed method can be used in combination with feature-based expression classification methods in large data sets for improving expression classification accuracies.

\end{abstract}

\section{Introduction}

Automated detection of facial expressions has gained significant importance in the recent years especially with regards to the design of real-time security surveillance systems, internet-based social networking applications \cite{one} and human computer interaction systems \cite{Mehdi}. The primary challenges for automated facial expression detection include variations introduced by pose, lighting, distortions, expression and occlusions. While image filtering techniques aid equalization of lightening and distortions, Eigen-value decomposition of faces (Eigen-faces) followed by Isomap clustering have been well-known to cluster variations in pose \cite{two}. Additionally, several supervised classification algorithms and publicly available data bases \cite{three} have shown significant success in classifying facial features, skin texture and basic expressions such as fear, sadness, happiness, anger, disgust, surprise. Most of the existing texture-based expression detection algorithms \cite{three} rely heavily on facial feature extraction and classifier training, and thereby incur significant computational complexity in the training phase. In this work, we propose a novel network-based clustering algorithm that is capable of separating the marginally classifiable expression faces from the easily classifiable ones. This method has two-fold advantages. First, this method can be used to reduce the overall computational time complexity for facial expression detection in a particular test data base of faces by subjecting only the faces with marginally classifiable expressions to complex feature-based classification. Second, the network-based metrics can be used to detect the most significant faces in the training data that are vital for feature-based expression classification tasks. Such network-based identification of most significant training image set has not been done in existing works so far. Identification of the most significant training faces can improve the existing accuracies in facial expression classification on facial test data sets. 

The existing facial expression detection algorithms can be broadly categorized into two categories: holistic methods \cite{survey} that focus on features of the full face, and geometric methods that depend on important parts of the face such as eye lids, eye brows, lips, nose etc. for expression detection \cite{ekman} \cite{ekman2}. The first category of methods focus on pre-defined template matching (active shape models)\cite{reff} or extraction of Eigen-face descriptors followed by clustering using neural networks \cite{eigenfaces} \cite{agarwal}, support vector machines \cite{SVM}, Naive-Bayes or Hidden Markov models \cite{reff}. The second category of methods rely on the extraction of gray-scale and color features corresponding to facial features (group of edges) \cite{reff}, texture, and changes in eye-lids, eye-brows, nose, lips, wrinkles and bulges using local binary patterns (LBP) \cite{ojala}, optical flow \cite{opticalflow} and pyramid extension of the histogram of gradient (PHOG) descriptors \cite{bosch} \cite{dalal}. For various classifier and training data sets, existing expression classification accuracies typically range between 50-95\% \cite{reff} \cite{Mehdi} while computation time can range from 4.5 seconds to a few hours. The proposed method aims at significantly reducing the computation time and improving expression classification accuracies in data bases with a large number of faces. In this work, two classification tasks are performed that include classification of images with facial occlusions and classification of faces with happy emotion, respectively. Unsupervised classification requires only 2 training images for cluster identification with a run-time of less than 1 second per image in a 2.6 GHz 2GB RAM Laptop.

\section{Proposed Method}
\begin{figure}[ht]
\begin{center}
\includegraphics[width = 3.0in, keepaspectratio=true]{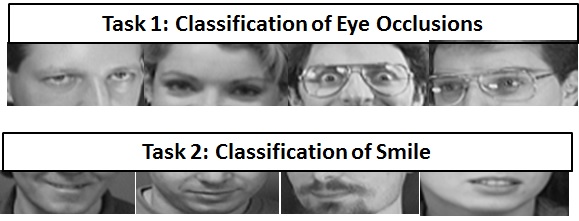}
\caption{The two proposed classification tasks. Automated facial patches isolate the regions of interest in faces and reduce the effects of pose and lighting variations.}     \label{tasks}
\end{center}
\end{figure}

In this work, we focus on two specific binary facial expression classification tasks. The first task involves classification of faces that have occlusions in the eye region, such as glasses, from faces without occlusions. The second task involves classification of faces with a smile from the non-smiling faces. Both tasks are challenging due to variations in pose and lighting angles. Some examples of the two binary classification tasks are shown in Figure \ref{tasks}. 

The data set used to analyze the expression classification performance are taken from the AT\&T Cambridge Laboratories data base \cite{att}, which contains 400 facial images of dimension [112x92] pixels each, with varying expression, pose and lighting angles from 40 subjects with 10 images per subject. For our analysis we manually annotate the expressions in 80 facial images corresponding to the first and tenth image per subject for the 40 subjects. For classification Task 1 the faces with glasses are assigned class label 1 and for the faces without glasses the class label is 0. For classification Task 2 the smiling faces are assigned class label 1 and for the faces without a smile the class label is 0. To reduce the computational complexity, each facial image `$I$' is resized to [90x90] pixels.

\begin{figure}[ht]
\begin{center}
\includegraphics[width = 2.35in, keepaspectratio=true]{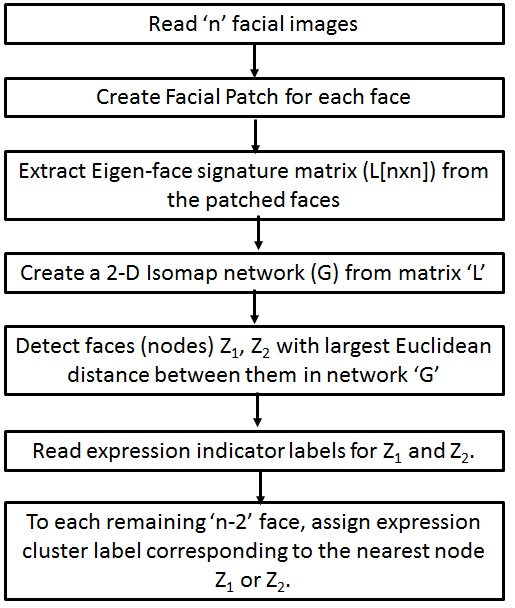}
\caption{Steps in the proposed algorithm.}     \label{flow}
\end{center}
\end{figure}

First, face patches corresponding to the eye region ($R_{eye}$), indicative of occlusions due to glasses, and the mouth region ($R_{mouth}$), indicative of smile are created. Next, Eigen-faces corresponding to the patched faces are extracted and a signature matrix of all the patched faces is created. For a set of $n$ faces, the Eigen-face signature matrix has dimensionality of [nxn]. Isomaps are then used to realize a 2-dimensional network from the facial signature matrix. Two nodes/faces in the network that have maximum Euclidean distance between them are detected as cluster identifiers followed by minimum-distance clustering of all remaining faces. The steps in the proposed algorithm are shown in Figure \ref{flow}. Each step is described below.

\subsection{Facial Patch Creation}
\begin{figure}[ht]
\begin{center}
\includegraphics[width = 3.5in, keepaspectratio=true]{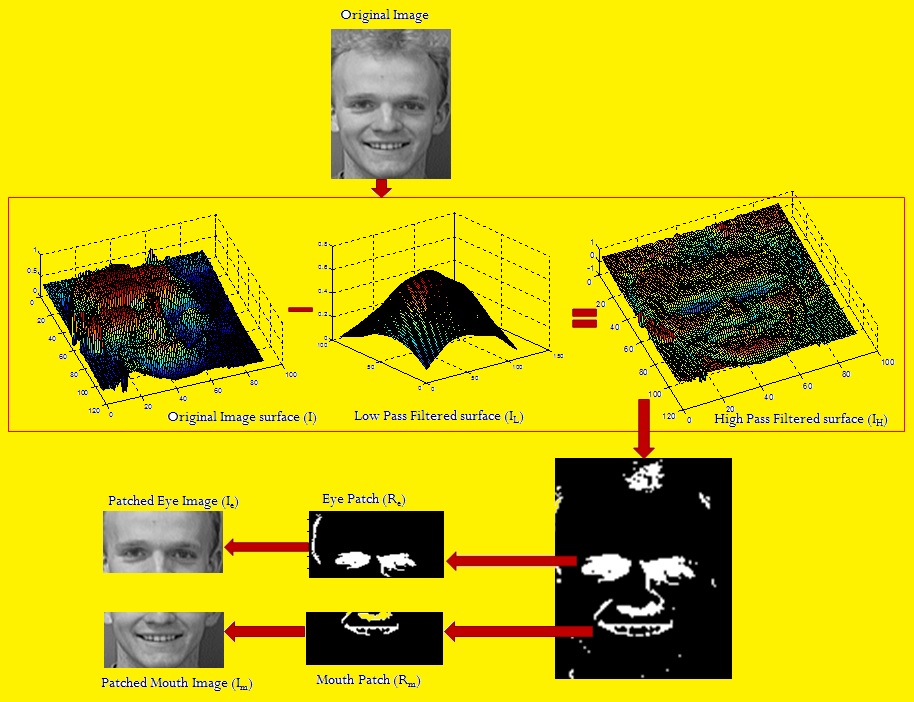}
\caption{Extraction of high pass filtered regions of interest and face patches corresponding to the eye and mouth region in each face, respectively.}     \label{Method}
\end{center}
\end{figure}

To generate guided patches for smiling expression and facial occlusion identification, each facial image is thresholded to generate a foreground mask followed by high-pass filtering to extract several regions of interest ($R$). 3 features are evaluated for each of the regions in `$R$’ namely, major axis length ($\rho(R)$), minor axis length ($\psi(R)$) and angular orientation ($\theta(R)$). The eye region ($R_{eye}$), indicative of occlusions due to glasses, contains high-pass filtered regions that are partially elliptical ($\frac{\rho(R)}{\psi(R)}<5$) and almost horizontal ($\arg\min(\theta(R))$) as shown in (\ref{pat}). The mouth region ($R_{mouth}$) on the other hand is elongated and narrow ($\frac{\rho(R)}{\psi(R)}>6$) and almost horizontal ($\arg\min(\theta(R))$) as shown in (2).

\begin{eqnarray}\label{pat}
R_1=2.5<\frac{\rho(R)}{\psi(R)}<5, R_{eye}=\arg\min_{R_1}\theta(R_1)\\
R_2=\frac{\rho(R)}{\psi(R)}>6, R_{mouth}=\arg\min_{R_2}\theta(R_2)
\end{eqnarray}
Finally, two face patches are created starting from the centroid of the region in `$R_{eye}$' and `$R_{mouth}$', respectively, and extending 15 pixels above and below the centroid with length similar to that of the original image. These patches applied to image `$I$' results in patched image of the eye ($I_e$) and mouth ($I_m$) as shown in Figure \ref{Method}, respectively. 

\subsection{Eigen-Face Generation}
For each patched image `$I_e, I_m$', the Karhunen-Loeve expansion \cite{four} is applied to find vectors that best represent the distribution of face images $\left\{I_{e1},I_{e2}...I_{en}\right\}, \left\{I_{m1},I_{m2}...I_{mn}\right\},n=80$. The steps of Eigen-face generation for the patched eye images ($I_e$) in Task 1 are shown below. Similar steps are followed for the patched mouth images ($I_m$) in Task 2.
The mean facial image is computed as the $0th$ Eigen-vector using (\ref{eq1}). The differences of each face from the average face are then computed using (4). It is noteworthy that the dimensionality of each resized face vector $\phi_i$ is [1x$q$], where `$q=90$x$90=8100$'.
 
\begin{eqnarray}\label{eq1}
\mu_{I_e}=\frac{1}{n} \sum_{i=1}^{n}I_{ei}\\
\forall i=1:n, \phi_i=I_{ei}-\mu_{I_e}
\end{eqnarray}

Each difference image is then subjected to principal component analysis (PCA) to find a set of `$n$' orthonormal vectors `$\nu_j|_{j=1}^{j=n}$', which best describe the distribution of the data set as shown in (\ref{eq2}).

\begin{eqnarray}\label{eq2}
\lambda_j=\frac{1}{n}\sum_{i=1}^{n}(\nu_j^T \phi_i)^2 {\bf is~maximum~such~that}\\\nonumber
\nu_l\nu_j=1, ~~~~~~~{\bf if~l=j},\\\nonumber
=0, {\bf otherwise}.
\end{eqnarray}

In (\ref{eq2}), $\nu_j$ and $\lambda_j$ are Eigen-vectors and Eigen-values of the covariance matrix in (\ref{eq3}).

\begin{equation}\label{eq3}
C_{ov}=\frac{1}{n}\sum_{i=1}^{n} \phi_i \phi_i^T=A.A^T,{\bf where,~}A=[\phi_1, \phi_2,.....\phi_n].
\end{equation}

Now, the real symmetric covariance matrix $C_{ov}$ has dimensions [$q$x$q$], and determination of `$q$' Eigen-vectors is an intractable operation for large image sizes. Thus, the computationally feasible solution for Eigen-vector determination in (\ref{eq2}) is to correlate the Eigen-vectors of $A^{T}.A$ with dimensionality [nxn] to that of $C_{ov}$ as shown in (\ref{eq4}).

\begin{eqnarray}\label{eq4}
{\bf If~}A^T.A. v_i=\lambda'_i .v_i,{\bf~ then,~}A.A^T .A. v_i=\lambda'_i .A.v_i\\ \nonumber
=>{\bf A.v_i~are~Eigen-vectors~of~C_{ov}=A.A^T}.
\end{eqnarray}

\begin{figure}[ht]
\begin{center}
\includegraphics[width = 3.25in, keepaspectratio=true]{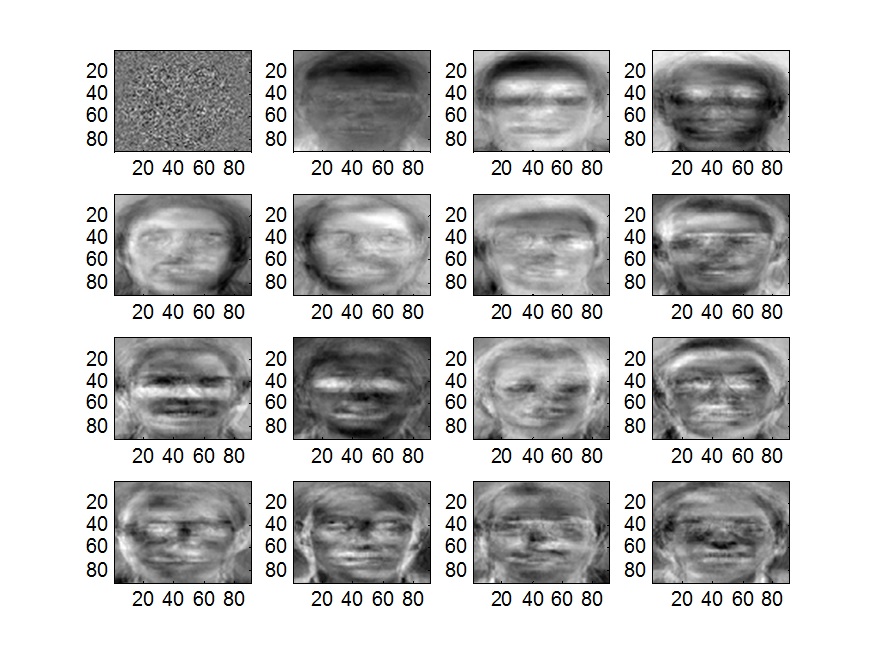}
\caption{Example of Eigen-Face generation with full facial images `$I$'. The top-left image represent the averaged image or the 0th Eigen-vector. The 1st to 15th Eigen-vectors of the first image from the data base are shown thereafter.}     \label{eigen}
\end{center}
\end{figure}

From this analysis, we construct matrix `$L=A^T.A$' of dimension [nxn], where $L_{l,q}=\phi_l^T.\phi_q$. `$n$' Eigen-vectors `$v_i$' of matrix `$L$' determine the linear contributions of  `$n$' faces to form Eigen-faces $u_i=\sum_{i=1}^{n}v_{ij}\phi_j$. The impact of Eigen-face generation is shown in Figure \ref{eigen}, where for the whole facial images `$I$', the $0th$ Eigen-vector followed by top 15 Eigen-vectors for an image are shown. The matrix $L=A^T.A$ represents the signature of each face in terms of an `$n$' dimensional vector. Next, Isomaps are generated using this matrix `$L$' for lower dimension embedding by multi-dimensional scaling \cite{Isomap}.
             
\subsection{Facial Network Clustering: Isomaps}
Isomaps have been used to find lower-dimensional manifolds from high-dimensional input data points for clustering faces based on imaging angles and lighting effects \cite{Isomap}\cite{Isomap1}. In this work, we apply the same principle to analyze the impact of low dimension embedding on facial expression clustering.

Using the [$n$x$n$] Eigen-face signature matrix `$L$', Isomaps are used to realize an unweighted network `$G$', where each facial image `$i$', where $i=[1,2...n]$ is connected to `$k$' nearest Euclidean neighbors. The network `$G=(Y,E)$', where $Y_i|_{i=1}^{n}$ represents the n-dimensional signature of each Eigen-face as a node, and `$E$' represents the connectivity matrix. It is noteworthy that if `$k$' is very large, too many connections destroy the clustering pattern, while a small `k' creates a sparse network that lacks clustering properties. The Euclidean distance between nodes $Y_i$ and $Y_j$ are represented as $D_{i,j}|_{i,j=1}^{n}$.

For the two binary expression clustering tasks at hand two faces are identified as representatives of cluster 0 ($Z_1$) and cluster 1 ($Z_2$), respectively. The two faces (nodes) that have the largest Euclidean distance between them are selected as the respective cluster representatives using (\ref{eq5}). The manually annotated expression class labels of faces $Z_1$ ($C_{Z_1}$) and $Z_2$ ($C_{Z_2}$) corresponding to the classification tasks are then read. These two faces become the training data. For every remaining face, the Euclidean distance of each face from $Z_1$ and $Z_2$ are computed, followed by assignment of the class label that is at the shortest distance from each face as shown in (9). The assigned class label are [0,1] due to the binary classification tasks.

\begin{eqnarray}\label{eq5}
[Z_1,Z_2]=\arg\max_{i,j={1,2..n}}D_{i,j}\\
\forall i=[1,2...n-2], C_i=C_{Z_1},{\bf If } D_{i,Z_1}<D_{i,Z_2}\\ \nonumber
=C_{Z_2},{\bf If } D_{i,Z_1}>D_{i,Z_2}\\\nonumber
\end{eqnarray}

\section{Experiments and Results}
The performance of the proposed patched Eigen-face based Isomap clustering for facial occlusion and happiness expression classification are evaluated using two experiments. In the first experiment, the best classification metrics obtained using patched Eigen faces and full Eigen-faces are comparatively analyzed. For both classification tasks, the number of faces with manually annotated expression class label 1 that are correctly classified are true positives ($tp$), faces with manual expression class label 0 that are correctly classified are true negatives ($tn$). Faces that are actually manually annotated as class label 1 but misclassified as class 0 are false negatives ($fn$) and faces with actual manual class label 0 but misclassified as 0 are false positives ($fp$). Classification performance metrics are computed as sensitivity (SEN), also known as recall, specificity (SPEC) and accuracy (ACC) are computed using (\ref{eqn4}) .

\begin{eqnarray}\label{eqn4}
SEN=\frac{tp}{tp+tn},SPEC=\frac{tn}{tn+fp},ACC=\frac{tp+tn}{n-2},\\\nonumber
{\bf where, }~~tp+tn+fp+fn=n-2.\\ \nonumber
\end{eqnarray}

In the second experiment, the 2-d patched face network is analyzed to detect the faces that have most discriminating characteristics for expression classification.

\subsection{Performance of Expression Classification}
In Table \ref{denoise_tab}, we observe that for low nearest neighbor parameters ($k=3, 5, 7$), the Isomaps demonstrate sparsely connected clustering patterns that have high classification characteristics than for higher values of `$k$'. Also, we observe that patched Eigen-face networks have comparable classification performance for occlusions when compared to  full Eigen-face networks. However, in Table \ref{denoise_tab} the proposed patched Eigen-face based networks are shown to improve classification SEN, ACC and area under Receiver Operating Characteristic curve (AUC) over full Eigen-face based networks for both classification tasks. 

\begin{table}[ht]
\caption{Isomap clustering for occlusion and smile classification tasks with the proposed patched Eigen-faces in comparison with full Eigen-faces.}
\begin{tabular}{|c c c c c c|}
\hline
Method&SEN&SPEC&ACC&k&AUC\\\hline
Task 1:&Occlusion&&&&\\	\hline				
Full Faces&0.6896&0.745&0.725&5&0.7031\\\hline
Patched Faces&0.7586&0.6862&0.725&5&0.7245\\\hline \hline
Task 2:&Smile&&&&\\\hline				
Full Faces&0.1428&0.8667&0.55&3&0.5111\\ \hline
Patched Faces&0.75&0.5556&0.6625&7&0.6319\\ \hline
\end{tabular}
	\label{denoise_tab}
\end{table}

\subsection{Expression Network Analysis}
After all faces have been clustered using (8-9), the most significant faces that are central to the two expression clusters are detected using network centrality measures. High betweenness centrality ($B$) locates nodes in a network that serve as bridge nodes connecting two dense clusters \cite{thesis}. Thus, the two faces/nodes in the Isomap network that have the top two betweenness centralities ($B_1$, $B_2$) represent the faces that lie at the edge of the expression clusters as the marginally classifiable faces. 

\begin{figure*}[ht]
\begin{center}
\subfigure[]{\includegraphics[width = 2.85in, keepaspectratio=true]{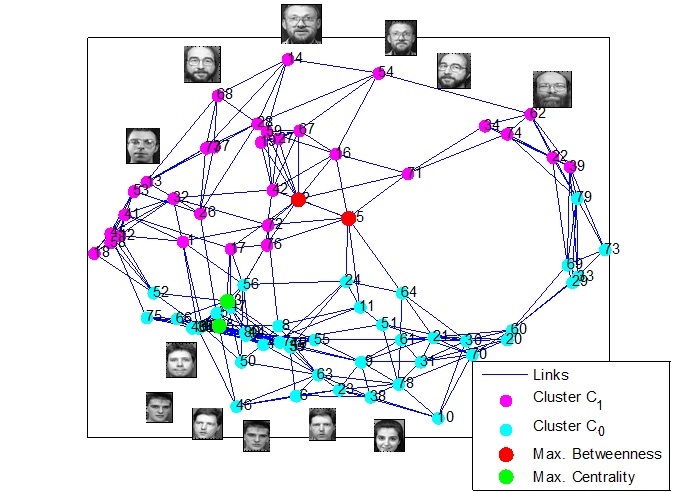}}
\subfigure[]{\includegraphics[width = 2.85in, keepaspectratio=true]{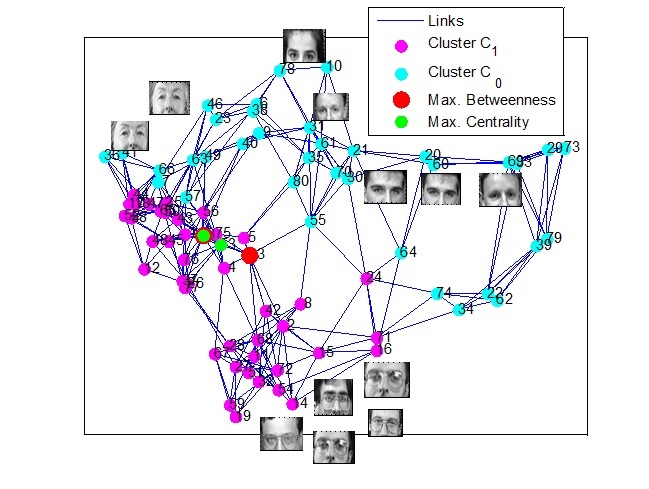}}
\subfigure[]{\includegraphics[width = 2.85in, keepaspectratio=true]{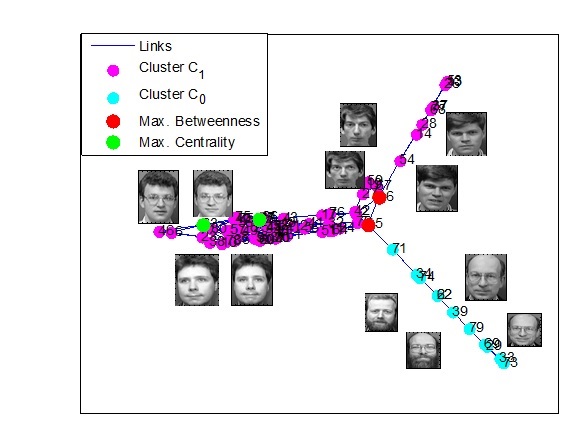}}
\subfigure[]{\includegraphics[width = 2.85in, keepaspectratio=true]{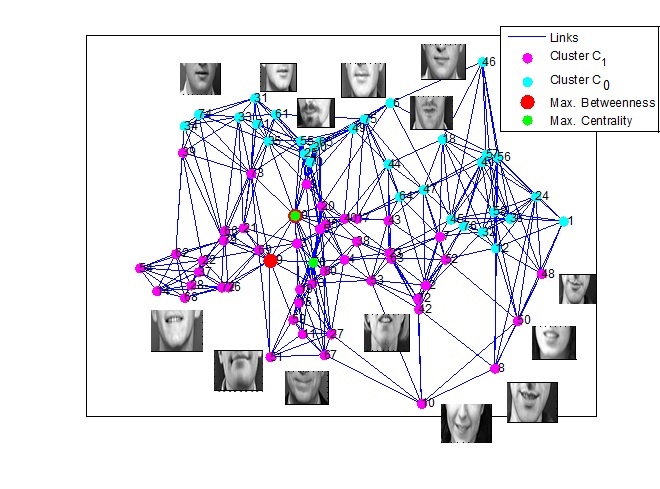}}
\caption{Isomap clustering for expression classification and detection faces with high centrality measures. (a) Full Eigen-Face clustering for Task 1. (b) Proposed patched Eigen-face clustering for Task 1. (c) Full Eigen-Face clusters for Task 2. (d) Patched Eigen-face clusters for Task 2. } \label{network}
\end{center}
\end{figure*}

High Eigen-centrality($EC$) is another measure that detects nodes in the network that are most centrally located. Thus, the faces with top two Eigen-centralities ($EC_1$, $EC_2$) represent the faces that are central to the two decision clusters. For Task 1 and Task 2 the decision clusters and the faces/nodes with top two centrality metrics for full face and proposed patched face network are shown in Figure \ref{network}.

\begin{figure}[ht]
\begin{center}
\includegraphics[width = 3.2in, keepaspectratio=true]{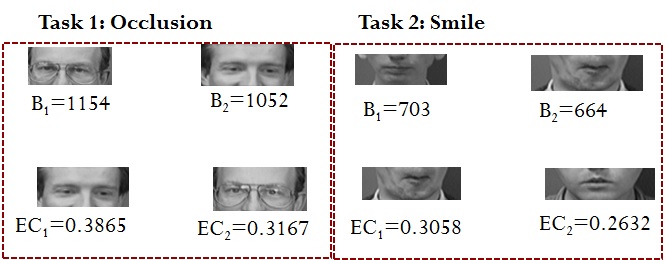}
\caption{Faces with top two $B$ and $EC$ measures for Task 1 and Task 2.} \label{central}
\end{center}
\end{figure}

In Figure \ref{network} we observe that for full Eigen-face networks, the nodes/faces with high $B$ are different from faces with high $EC$, however for the patched face networks, certain faces concurrently have high $B$ and $EC$. In Task 1 and Task 2, the most central patched faces are shown in Figure \ref{central}. From this observation we infer that patched faces with high centralities are central to the expression clusters and they also serve as bridge node connections to the other cluster. Thus, identification of these faces/nodes with high centrality measure and using them as training data can further improve the feature-based expression classification performances. 

\begin{figure}[ht]
\begin{center}
\subfigure[]{\includegraphics[width = 2.5in, keepaspectratio=true]{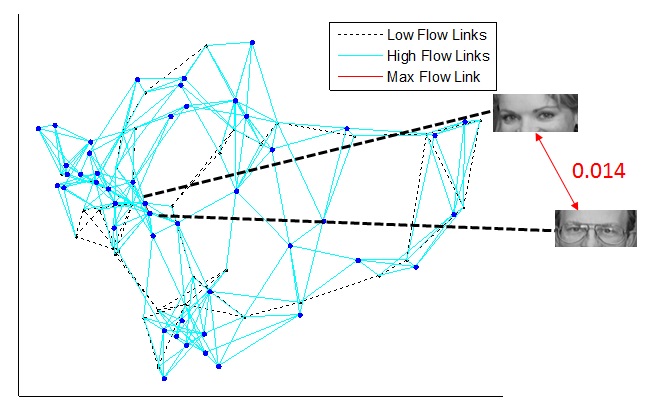}}
\subfigure[]{\includegraphics[width = 2.6in, keepaspectratio=true]{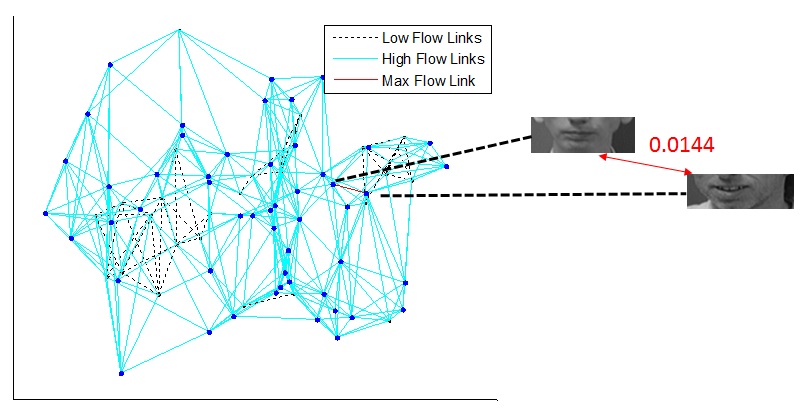}}
\caption{Max-flow-min-cut applied to isolate nodes/faces with most information flow through them. The maximum flow as a fraction of the total information flow through the network is shown for (a) Task 1, (b) Task 2.} \label{cut}
\end{center}
\end{figure}

Additionally, we identify other significant patched faces for training purposes by applying repetitive max-flow-min-cut strategy to separate the links with high information flow through them, from the links with low flow. The nodes/faces at either ends of the link with maximum information flow through it are indicative of the patched faces with most information. In Figure \ref{cut}(a) we observe that for Task 1, one instance of maximum information flow through the Isomap network occurs between a non-occluded female eye image and an occluded male eye image. In Figure \ref{cut}(b), for Task 2, another instance of maximum information flow through the Isomap network occurs between a non-smiling and a partially smiling facial image. Thus, several additional instances of max-flow-min-cut can identify the faces that are more significant than the others for feature-based expression classification tasks.

\section{Conclusions}
In this work we propose an automated facial patch creation method to isolate certain regions of a face regardless of the pose and lighting conditions followed by Eigen-face decomposition and Isomap clustering for classification of facial expressions. We observe that patched Eigen-face Isomap networks created with low neighborhood parameter ($k$) values have higher sensitivity than full Eigen-face networks for facial occlusion and smile classification tasks. Additionally, network-based centrality and information flow in networks can be used as measures to detect the most significant subset of faces for feature-based expression classification tasks. The proposed method requires an average of 0.25 seconds per image for generating automated facial masks, 0.75 seconds for Eigen-face decomposition and Isomap network creation followed by 0.1s for clustering the faces/nodes for each classification task. 

Future efforts will be directed towards using the proposed method as a first pass for expression classification tasks followed by feature-based expression on the marginally classifiable faces using the most significant faces detected from the proposed method as training data set. Future works can also investigate the performance of the proposed method on other expression classification tasks such as fear, disgust and anger in images with multiple faces.

\bibliographystyle{named}
\bibliography{ijcai13,references}

\begin{thebibliography}{}

\bibitem[\protect\citeauthoryear{Agarwal \bgroup \em et al.\egroup
  }{2010}]{agarwal}
Mayank Agarwal, Nikunj Jain, Mr~Manish Kumar, and Himanshu Agrawal.
\newblock Face recognition using eigen faces and artificial neural network.
\newblock {\em International Journal of Computer Theory and Engineering},
  2(4):1793--8201, 2010.

\bibitem[\protect\citeauthoryear{Al-modwahi \bgroup \em et al.\egroup
  }{2012}]{one}
Ashraf Abbas~M Al-modwahi, Onkemetse Sebetela, Lefoko~Nehemiah Batleng, Behrang
  Parhizkar, and Arash~Habibi Lashkari.
\newblock Facial expression recognition intelligent security system for real
  time surveillance.
\newblock In {\em Proc. of World Congress in Computer Science, Computer
  Engineering, and Applied Computing}, 2012.

\bibitem[\protect\citeauthoryear{Bosch \bgroup \em et al.\egroup
  }{2007}]{bosch}
Anna Bosch, Andrew Zisserman, and Xavier Munoz.
\newblock Representing shape with a spatial pyramid kernel.
\newblock In {\em Proceedings of the 6th ACM international conference on Image
  and video retrieval}, pages 401--408. ACM, 2007.

\bibitem[\protect\citeauthoryear{Chavan and Kulkarni}{2013}]{three}
Umesh~Balkrishna Chavan and Dinesh~B Kulkarni.
\newblock Facial expression recognition-review.
\newblock {\em International Journal of Latest Trends in Engineering and
  Technology (IJLTET)}, 3(1):237--243, 2013.

\bibitem[\protect\citeauthoryear{Dalal and Triggs}{2005}]{dalal}
Navneet Dalal and Bill Triggs.
\newblock Histograms of oriented gradients for human detection.
\newblock In {\em Computer Vision and Pattern Recognition, 2005. CVPR 2005.
  IEEE Computer Society Conference on}, volume~1, pages 886--893. IEEE, 2005.

\bibitem[\protect\citeauthoryear{Dehkordi and Haddadnia}{2010}]{opticalflow}
Behnam~Kabirian Dehkordi and Javad Haddadnia.
\newblock Facial expression recognition in video sequence images by using
  optical flow.
\newblock In {\em Signal Processing Systems (ICSPS), 2010 2nd International
  Conference on}, volume~1, pages V1--727. IEEE, 2010.

\bibitem[\protect\citeauthoryear{Ekman and Friesen}{1971}]{ekman2}
Paul Ekman and Wallace~V Friesen.
\newblock Constants across cultures in the face and emotion.
\newblock {\em Journal of personality and social psychology}, 17(2):124, 1971.

\bibitem[\protect\citeauthoryear{Ekman and Rosenberg}{1997}]{ekman}
Paul Ekman and Erika~L Rosenberg.
\newblock {\em What the face reveals: Basic and applied studies of spontaneous
  expression using the Facial Action Coding System (FACS)}.
\newblock Oxford University Press, 1997.

\bibitem[\protect\citeauthoryear{Hemalatha and Sumathi}{2014}]{reff}
G~Hemalatha and CP~Sumathi.
\newblock A study of techniques for facial detection and expression
  classification.
\newblock {\em International Journal of Computer Science \& Engineering Survey
  (IJCSES) Vol}, 5, 2014.

\bibitem[\protect\citeauthoryear{Kirby and Sirovich}{1990}]{four}
Michael Kirby and Lawrence Sirovich.
\newblock Application of the karhunen-loeve procedure for the characterization
  of human faces.
\newblock {\em IEEE Transactions on Pattern Analysis and Machine Intelligence},
  12(1):103--108, 1990.

\bibitem[\protect\citeauthoryear{Laboratories}{2002}]{att}
At\&T~Cambridge Laboratories.
\newblock The database of faces.
\newblock {\em http://www.cl.cam.ac.uk/research/dtg/attarchive
  /facedatabase.html}, 2002.

\bibitem[\protect\citeauthoryear{Lajevardi and Wu}{2012}]{Mehdi}
Seyed~Mehdi Lajevardi and Hong~Ren Wu.
\newblock Facial expression recognition in perceptual color space.
\newblock {\em Image Processing, IEEE Transactions on}, 21(8):3721--3733, 2012.

\bibitem[\protect\citeauthoryear{Li \bgroup \em et al.\egroup }{2005}]{two}
Rui-Fan Li, Hong-Wei Hao, Xu~yan Tu, and Cong Wang.
\newblock Face recognition using kfd-isomap.
\newblock In {\em Proceedings of 2005 International Conference on Machine
  Learning and Cybernetics, 2005.}, volume~7, pages 4544--4548 Vol. 7, Aug
  2005.

\bibitem[\protect\citeauthoryear{Lonare and Jain}{2013}]{survey}
Ashish Lonare and Shweta~V Jain.
\newblock A survey on facial expression analysis for emotion recognition.
\newblock {\em International Journal of Advanced Research in Computer and
  Communication Engineering}, 2(12), 2013.

\bibitem[\protect\citeauthoryear{Ojala \bgroup \em et al.\egroup
  }{2002}]{ojala}
Timo Ojala, Matti Pietik{\"a}inen, and Topi M{\"a}enp{\"a}{\"a}.
\newblock Multiresolution gray-scale and rotation invariant texture
  classification with local binary patterns.
\newblock {\em Pattern Analysis and Machine Intelligence, IEEE Transactions
  on}, 24(7):971--987, 2002.

\bibitem[\protect\citeauthoryear{Punitha and Geetha}{2013}]{SVM}
A~Punitha and M~Kalaiselvi Geetha.
\newblock Texture based emotion recognition from facial expressions using
  support vector machine.
\newblock {\em algorithms (eg Hidden Markov Models (HMMs)}, 1:6, 2013.

\bibitem[\protect\citeauthoryear{Roychowdhury}{2010}]{thesis}
Sohini Roychowdhury.
\newblock {\em Mathematical models for prediction and optimal mitigation of
  epidemics}.
\newblock PhD thesis, Kansas State University, 2010.

\bibitem[\protect\citeauthoryear{Tenenbaum \bgroup \em et al.\egroup
  }{2000}]{Isomap1}
Joshua~B Tenenbaum, Vin De~Silva, and John~C Langford.
\newblock A global geometric framework for nonlinear dimensionality reduction.
\newblock {\em Science}, 290(5500):2319--2323, 2000.

\bibitem[\protect\citeauthoryear{Turk and Pentland}{1991}]{eigenfaces}
Matthew Turk and Alex Pentland.
\newblock Eigenfaces for recognition.
\newblock {\em Journal of cognitive neuroscience}, 3(1):71--86, 1991.

\bibitem[\protect\citeauthoryear{Yang}{2002}]{Isomap}
Ming-Hsuan Yang.
\newblock Extended isomap for pattern classification.
\newblock In {\em AAAI/IAAI}, pages 224--229, 2002.

\end{thebibliography}

\end{document}